\documentclass[lettersize,journal]{IEEEtran}
\usepackage{amsmath,amsfonts}
\usepackage{algorithmic}
\usepackage{algorithm}
\usepackage{array}
\usepackage[caption=false,font=normalsize,labelfont=sf,textfont=sf]{subfig}
\usepackage{textcomp}
\usepackage{stfloats}
\usepackage{url}
\usepackage{verbatim}
\usepackage{graphicx}
\usepackage{cite}
\begin{document}

\title{Selective Task offloading for Maximum Inference Accuracy and Energy efficient Real Time IoT Sensing Systems}

\author{Abdelkarim Ben Sada, Amar Khelloufi, Abdenacer Naouri,  Huansheng Ning and Sahraoui Dhelim
        
\thanks{Abdelkarim Ben Sada, Abdenacer Naouri, Huansheng Ning, Amar Khelloufi  are with the University of Science and Technology Beijing, Beijing, China}
\thanks{Sahraoui Dhelim is with the School of Computer Science, University College Dublin, Ireland}
\thanks{Corresponding author: Sahraoui Dhelim (sahraoui.dhelim@ucd.ie).}
\thanks{Manuscript received February 22, 2024; revised February 22, 20124.}
}

\markboth{Journal of \LaTeX\ Class Files,~Vol.~14, No.~8, July~2023}%
{Shell \MakeLowercase{\textit{et al.}}: A Sample Article Using IEEEtran.cls for IEEE Journals}


\maketitle

\begin{abstract}
The recent advancements in small-size inference models facilitated AI deployment on the edge. However, the limited resource nature of edge devices poses new challenges especially for real-time applications. Deploying multiple inference models (or a single tunable model) varying in size and therefore accuracy and power consumption, in addition to an edge server inference model, can offer a dynamic system in which the allocation of inference models to inference jobs is performed according to the current resource conditions. Therefore, in this work, we tackle the problem of selectively allocating inference models to jobs or offloading them to the edge server to maximize inference accuracy under time and energy constraints. This problem is shown to be an instance of the unbounded multidimensional knapsack problem which is considered a strongly NP-hard problem. We propose a lightweight hybrid genetic algorithm (LGSTO) to solve this problem. We introduce a termination condition and neighborhood exploration techniques for faster evolution of populations. We compare LGSTO with the Naive and Dynamic programming solutions. In addition to classic genetic algorithms using different reproduction methods including NSGA-II, and finally we compare to other evolutionary methods such as Particle swarm optimization (PSO) and Ant colony optimization (ACO). Experiment results show that LGSTO performed 3 times faster than the fastest comparable schemes while producing schedules with higher average accuracy.
\end{abstract}


\section{Introduction}
AI has become an integral part of our daily life and it continues to advance everyday. The introduction of small-size AI models which can be deployed on resource-constrained edge devices has opened the door for low-latency applications in IoT. Various applications of these local models include real-time image/video recognition and augmented reality in smartphones. This enables features such as object detection, facial recognition, and AR overlays without the need to upload data to remote servers, enhancing power efficiency, privacy and latency. AI-powered applications such video editing apps using smartphones can benefit from local models to offer real-time previews of edited frames.\cite{singh2023edge}

Numerous off-the-shelf advanced small models have been recently developed such as the large language model Gemini-nano on smartphones by Google. \cite{team2023gemini} These models offer on-device inference capabilities with reasonable runtimes and accuracy which allows for low latency services and efficient energy consumption compared to relying on cloud services. However, not all inference tasks can be performed locally while conserving energy and offering real-time results. Therefore, a balance between performing tasks locally on edge devices or offloading them to nearby edge servers is required to achieve optimal performance.\cite{xu2020energy}

Dedicated AI hardware embedded in modern processors is encouraging the deployment of real-time AI models in edge devices. By compromising on accuracy, multiple AI models with varying computation and storage requirements are being deployed at the edge. The size of the model is directly related to the inference time and accuracy, where larger models produce more accurate inference results while taking longer processing times. On the other hand, smaller models take less time to process data resulting in reduced accuracy. For some proposed models the accuracy can be adjusted using hyper-parameters. This allows the edge node to have multiple local inference models with varying accuracies.\cite{fresa2021offloading}

Edge computing aims to provide computation and storage closer to the edge devices reducing latency. By deploying edge servers nearby the edge nodes, resource intensive tasks can be offloaded to these servers for better response times. However, communication costs in terms of energy consumption and latency have to be considered specially for battery powered edge devices. \cite{li2019edge}

Task offloading for edge computing has been extensively studied in the literature for various application domains \cite{Abdenacer2023,fog_placement}. However, when it comes to offloading tasks regarding AI models where the accuracy, processing time and energy play an important role in improving the end user experience, less attention is given to the topic from the research community. 

\begin{figure}
\centering
\includegraphics[width=3.4in]{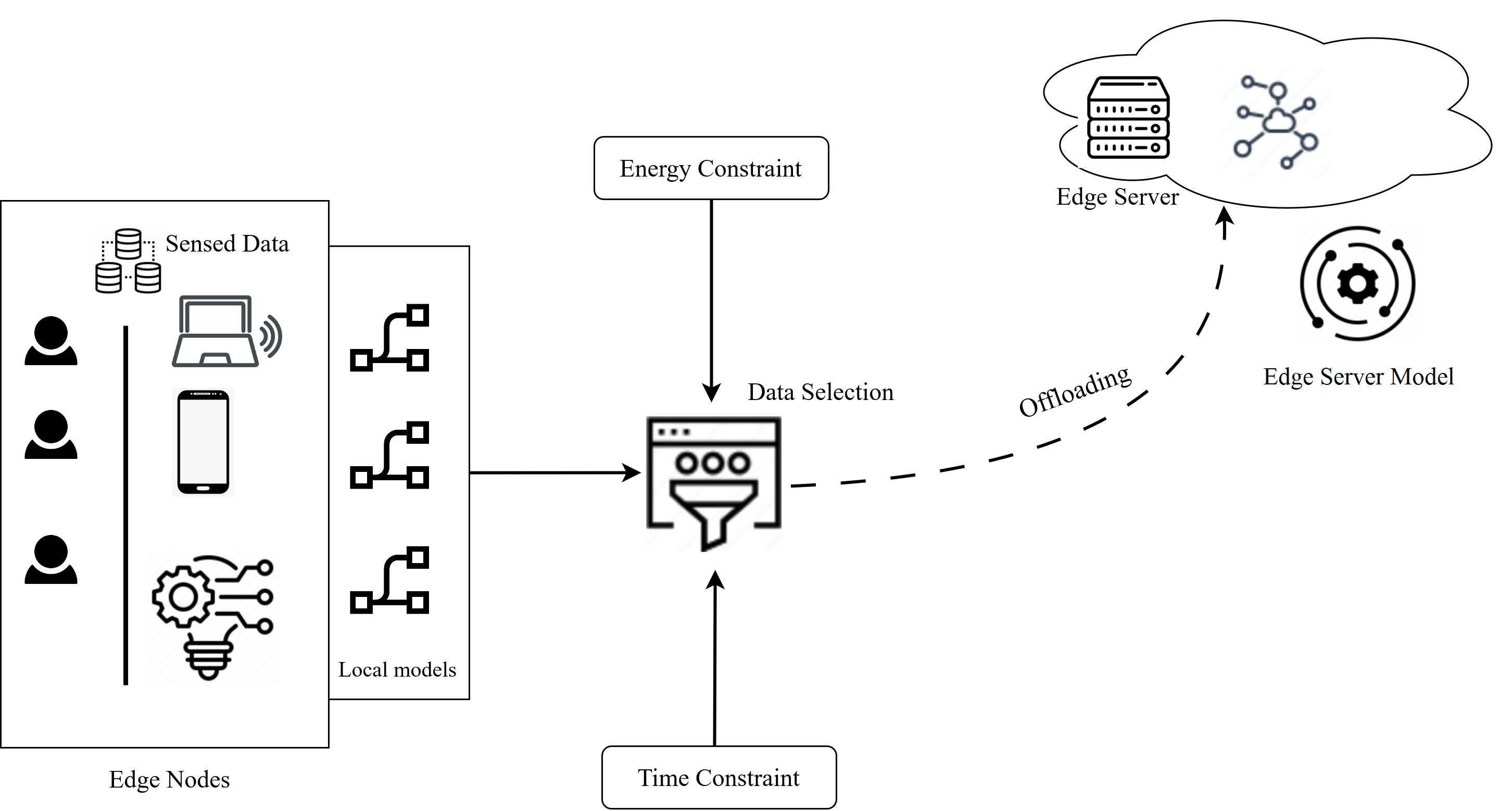}
\caption{Inference model selection between local models and the edge server.}   
\label{fig:sys_architecture}
\end{figure}

Considering the system in Fig.\ref{fig:sys_architecture}, every edge node is equipped with multiple inference models varying in size and thus accuracy and inference time. These models can be of the same type (e.g. DNN) with different internal structures or can be of completely different types (e.g. SVM, RNN). The edge server is equipped with a higher accuracy model with a relatively short inference time since it has more computing capacity and virtually unlimited power. 

In the context of edge computing devices constrained by limited energy and computing resources, the selection of appropriate inference models becomes a critical aspect for optimizing performance. Equipped with multiple inference models, these edge devices must navigate a delicate trade-off between accuracy, time efficiency, and energy consumption. The challenge lies in selecting the right model for a specific task that aligns with the device's computational capabilities and energy constraints. Striking a balance between accuracy and resource efficiency is imperative to ensure that inference tasks are executed within acceptable time frames and without compromising the device's energy budget. This demands intelligent decision-making mechanisms, possibly leveraging dynamic model selection algorithms, that consider the nature of the task, current system conditions, and the available models' computational demands. By dynamically adapting to the contextual needs of each inference job, edge devices can maximize their overall efficiency, providing accurate results while respecting strict time and energy limitations.\cite{fresa2021offloading,BenSada2023,context_siot}

The challenge of efficiently assigning inference jobs to local inference models or deciding whether to offload them to an edge server draws parallels with the Unbounded Multidimensional Knapsack problem (UMdKP), a variant of the classic knapsack problem. The UMdKP involves selecting items from an unbounded set, each with multiple dimensions (constraints), to maximize the overall value without exceeding the given constraints. Analogously, in the context of edge computing, the decision to allocate inference tasks locally or offload them to an edge server involves managing multiple constraints, such as time and energy, while optimizing accuracy. The UMdKP is known to be strongly NP-Hard, implying that finding an optimal solution within polynomial time is unlikely. Various solutions have been proposed in the literature, employing pseudo-polynomial time approaches like Dynamic Programming to tackle the UMdKP. In this research, we introduce a novel hybrid meta-heuristic method designed to address the inference scheduling problem under the constraints of time and energy while simultaneously maximizing inference accuracy. This approach aims to provide an efficient and effective solution to the complex decision-making process inherent in edge computing environments.\cite{cacchiani2022knapsack}

The main contributions of this paper can be summarized as follows:

\begin{itemize}
    \item We formulate the novel problem of multi-inference model selection under time and energy constraints where we draw similarity to the popular UMdKP and propose a lightweight hybrid genetic algorithm LGSTO to tackle this problem.
    \item  We introduce a termination condition and a neighborhood exploration mechanisms to accelerate the evolution process and converge to the best solution in as few generations as possible.
    \item We perform experiments using raspberry pi and an edge server on the Imagenet-mini dataset and we
    \item Compare the performance of LGSTO against the Brute-force solution optimized with memoization, in addition to Dynamic programming solution. For evolutionary schemes we compare against genetic algorithms using different reproduction methods such Gene pool, 1-point crossover and NSGA-II. In addition to Particle Swarm Optimization (PSO) and Ant Colony Optimization (ACO). We find that LGSTO is 70\% faster then the fastest comparable schemes while producing higher average accuracy.
\end{itemize}

The rest of this paper is organized as follows. Section \ref{sec:related_works} presents the related works and points out the research gap. In Section \ref{sec:sys_model} we describe the system model. In Section \ref{sec:lgsto} we propose LGSTO and explain the solution steps. In Section \ref{sec:exp_results} we present the experiment setup and results in addition to analysis of the obtained results. Finally, we conclude this work in Section \ref{sec:conclusion}.

\section{Related Works}
\label{sec:related_works}
In this section we divide the related works into two main sections. Firstly, we present works that try to solve the inference job offloading. Secondly, we review the recently proposed algorithms that attempt to solve the UMdKP using meta-heuristic methods. 

\subsection{Inference Job offloading}
Computational offloading in mobile edge devices is a very popular topic because of the numerous benefits it brings. However, the offloading of ML inference jobs has received little attention from the research community due to the novelty aspect of the topic. Therefore, in this section we will be focusing on the few recent works tackling this specific subject. 

In \cite{nikoloska2020data}, the authors proposed a data selection scheme for IoT devices in which an edge node can decide to offload data that would likely lead to inaccurate inference if processed locally and thus improving the overall accuracy of the whole system. Their data scheme performs the selection under a given energy constraint. The proposed scheme shows promising results, however, it does not consider the time constraint and thus renders it unsuitable for real time applications. In addition, their scheme only considers a single inference model in the edge node which does not offer many options in terms of maximizing accuracy. 

The authors of \cite{fresa2021offloading} studied the inference job offloading under a time constraint in a system where edge nodes are equipped with multiple inference models varying in size and accuracy in addition to an edge server. Their system leverages the fact that they have two parallel machines namely the edge device and the edge server. They proposed $AMR^2$, a scheduling scheme based on LP-Relaxation and rounding which considers all possible cases of scheduling two jobs between the edge device and the edge server. They relax the problem's constraint to take fractional values and then perform rounding to get the result. In the edge device they use dynamic programming to schedule the jobs. Their proposed scheme performed better than the greedy approach.  However, their system does not take into consideration the energy constraint of the edge node. Although, using the edge device's onboard inference models along with offloading tasks to the edge server can reduce the total inference time of the system, in the case of battery powered edge devices, it costs a significant  amount of valuable energy.

The works presented in this section offer important insights into the problem of data selection and offloading in the context of inference models for improving energy in the case of \cite{nikoloska2020data}, or time and accuracy in \cite{fresa2021offloading}. Therefore, in this work we try to cover the research gaps and build on top of the limitations left by these works by proposing an improved solution.

\subsection{Task offloading in edge computing}

The work presented by \cite{chen2021energy} addresses the challenges posed by the high computational costs of Deep Neural Networks (DNNs) in energy-constrained Internet-of-Things (IoT) devices. The authors propose a novel system energy consumption model that accounts for runtime, switching, and computing energy consumption on servers and IoT devices in cloud-edge environments. Their strategy, based on the Self-adaptive Particle Swarm Optimization algorithm using Genetic Algorithm operators (SPSO-GA), efficiently makes offloading decisions for DNN layers, reducing energy consumption and improving execution time. However, this work highlights the ongoing challenges in offloading DNN layers while considering both deadline constraints and energy consumption, particularly in dynamic cloud-edge environments.

In \cite{cozzolino2022nimbus}, the focus shifts to Augmented Reality (AR) applications, emphasizing the importance of smooth and immersive experiences. The authors introduce Nimbus, a task placement and offloading solution designed for multi-tier edge-cloud infrastructure. By offloading computationally intensive tasks from AR applications to GPU-powered edge devices, Nimbus significantly reduces task latency and energy consumption. The work underscores the necessity of optimizing offloading policies for load distribution across edge nodes. This study's contributions lie in benchmarking edge device performance, proposing a resource provisioning algorithm, and evaluating Nimbus against other solutions in terms of latency and energy efficiency.

The paper by \cite{xu2020energy} explores the intersection of Deep Neural Networks (DNNs) and mobile edge clouds in 5G-enabled environments. It addresses the challenge of offloading inference requests while considering continuously generated data streams and the energy-intensive nature of 5G base stations and cloudlets. The authors propose exact and approximate algorithms, including learning-based dynamic inference offloading methods, to minimize energy consumption and meet stringent application delay requirements. This work distinguishes itself by considering the unique characteristics of 5G-enabled Mobile Edge Clouds (MECs) and formulating the inference offloading problem to optimize energy efficiency while meeting application demands.

Authors in \cite{sacco2021resource} present RITMO, a distributed and adaptive task offloading algorithm, specifically designed for Unmanned Aerial Vehicles (UAVs) in edge computing applications. RITMO utilizes a regressor to predict future UAV task queues, facilitating proactive and energy-aware task assignment. This work addresses challenges in dynamically reassigning tasks in challenged network scenarios and outperforms existing solutions in terms of overall latency and energy consumption. Notably, RITMO introduces a forward-looking approach by incorporating resource usage prediction, setting it apart from centralized and distributed alternatives.
Authors in \cite{wu2023fast} introduces a novel optimization framework, IOPO, for task offloading in Intelligent Reflecting Surface (IRS)-assisted Multi-Access Edge Computing Systems operating in Terahertz communication networks. IOPO utilizes deep learning, specifically the Iterative Order-Preserving Policy Optimization, to generate energy-efficient task-offloading decisions rapidly. The study distinguishes itself by integrating IRS and UAVs into the MEC system within the unique context of Terahertz communication networks. IOPO's ability to handle complex problems and generate optimal offloading decisions quickly sets it apart from traditional numerical optimization methods. 

The study conducted by \cite{abdenacer2023task} concentrates on task-offloading scenarios for healthcare applications performed on smart wearable glasses. The authors explore the optimal conditions for offloading computationally intensive tasks to nearby devices, such as mobile devices or remote servers. A specific use case is presented in the context of airport security, where smart glasses are used to detect elevated body temperatures. The contributions of this work include presenting a two-tier edge infrastructure, evaluating performance limitations of wearable devices, and investigating conditions for effective task offloading. This study underscores the practicality and relevance of task offloading in healthcare settings.
In the realm of Mobile Edge Computing (MEC), efficient task offloading and resource allocation are pivotal to enhancing user experience by leveraging nearby mobile edge servers (MES) through wireless access networks. Similarly, Dhelim et al. studied selective computing in various scenarios, including selective image classification \cite{Chaib2022,ipunet}, selective traffic routing \cite{Aung2023,Aung2020,Zhang2018}, social networks content selection \cite{dhelim2021social,dhelim2022artificial,dhelim2022hybrid,Dhelim2020}, and IoT \cite{Wei2022,web3,aung2022blockchain,SahraouiDhelim2016}

The authors in \cite{li2020genetic} propose a Genetic Algorithm (GA)-based joint optimization approach for task offloading proportion, channel bandwidth, and MES computing resources. The focus lies on scenarios where certain computing tasks can be partially offloaded to MES, optimizing the completion time of user tasks within the constraints of wireless transmission and MES processing resources. Unlike existing literature that predominantly tackles complete offloading or energy considerations, this work introduces a novel perspective by addressing proportional offloading scenarios. The key contributions of this paper include solving the intricate problem of minimizing overall completion time in scenarios involving multiple mobile devices and a single edge server, particularly when tasks can be divided proportionally. The proposed GA-based joint optimization algorithm proves effective in achieving optimal user task offloading proportions and resource allocations. However, the study acknowledges its simplicity, focusing solely on user completion time and omitting considerations of energy consumption, true distances, and scenarios involving multiple edge servers and base stations, marking potential avenues for future research exploration. However, This study diverges from our proposed work by utilizing Genetic Algorithms for offloading without incorporating machine learning models, emphasizing accuracy, or considering energy efficiency aspects.

In summary, while existing works have made significant strides in exploring energy-efficient task offloading, latency reduction, and resource optimization, our contributions introduce crucial dimensions that set our research apart. Unlike prior studies, we specifically address the complex challenges of job scheduling by conducting a meticulous comparative analysis of algorithms for the multi-dimensional, multi-constraint unbounded knapsack problem. Moreover, our novel genetic algorithm for real-time inference job scheduling on constrained edge devices represents a unique solution catering to both time and energy constraints. By incorporating extensive experiments our work stands out for its comprehensive evaluation, emphasizing factors such as energy consumption and real-world practicality. These nuanced considerations distinguish our contributions, highlighting their novel perspectives in enhancing the efficiency of task offloading systems in dynamic and resource-constrained environments.

\section{System Model}
\label{sec:sys_model}
\begin{figure}[!t]
\centering
\includegraphics[width=3.4in]{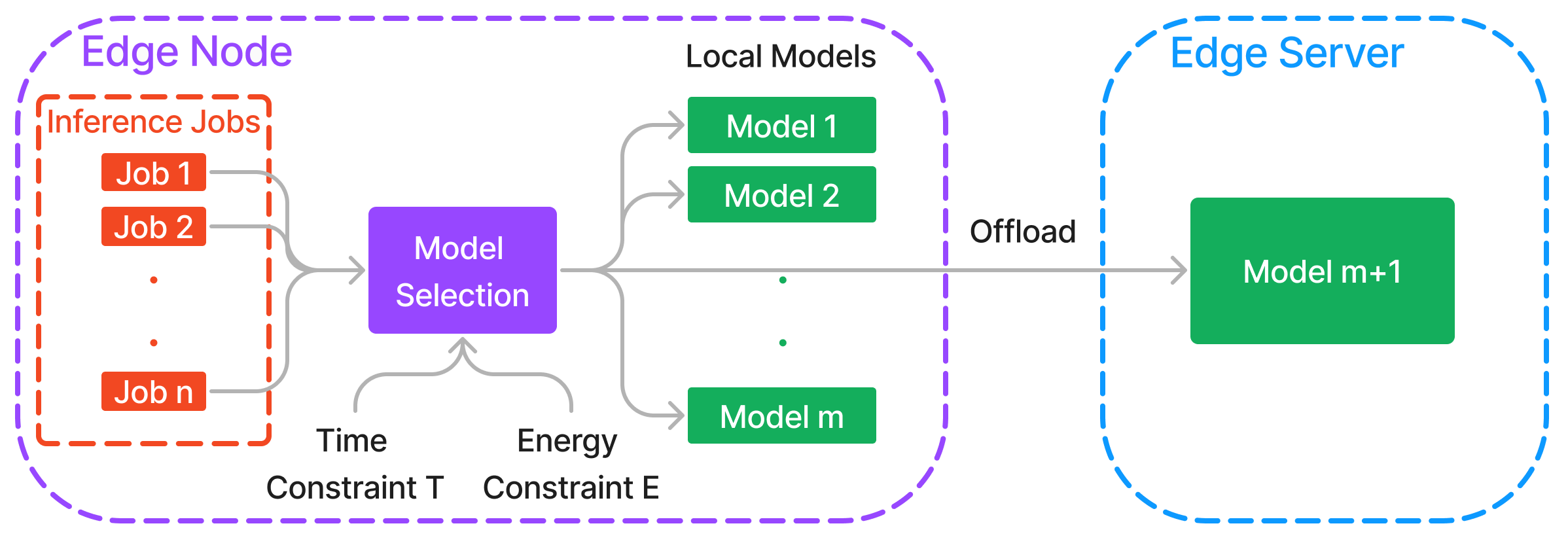}
\caption{System Model.}
\label{fig:sys_model}
\end{figure}

We consider a sensing system in which edge nodes receive a set of sensed data and perform inference on the data either locally using local inference models or by offloading the task to a nearby edge server (see Fig \ref{fig:sys_model}). The edge nodes must respect a given time and energy constraints to deliver the inference result. In this section, we present the modeling of the most important parts of the considered sensing system.

\subsection{Inference Models}
The edge nodes are equipped with $m$ local ML inference models, each model has an index $i$ where $i \in M$ and $M = \{1,...,m\}$. These models can be of the same type with different hyper-parameters or completely different types of models. The models vary in size and top-1 average accuracy thus having shorter or longer run-times depending on the size and accuracy. Since the real top-1 accuracy of each model for a given job is not known prior to performing the inference we use the average accuracy estimated over time by averaging historical real top-1 accuracies. The edge server inference model, given the index $m+1$, is considered to have the highest average top-1 accuracy since it is running on a more powerful hardware. let $a_i$ be the average top-1 accuracy of model $i$ and $a_{m+1}$ be the average top-1 accuracy of the edge server model $m+1$ where $a_{m+1} > a_i$.

An edge node receives a set of inference jobs (i.e. sensed data) at each time interval. Let $j$ be the index of an inference job where $j \in N$ and $N = \{1,...,n\}$.

\subsection{Processing Time}
Each inference model $i$ has an average inference time $it_i$ estimated by averaging the historical real inference times. We consider that the preprocessing time is part of $it_i$. Since the edge server is equipped with powerful hardware we consider that the inference time in the edge server $it_{m+1}$ to be constant where $it_{(m+1)} < it_{i}$.

Let $ot_j$ be the estimated time it takes to offload job $j$ to the edge server. We can estimate the offload time from the connection speed and the size of data to be offloaded which is given by:

$$ot_j = s_j / b$$

where $s_j$ is the size of job $j$ and $b$ is the bandwidth of the communication channel between the edge node and edge server. To simplify our work and keep the scope of this work focused on the selection problem, we consider the communication channel to be noise free. 

We define $t_{ij}$ as the time it takes to process job $j$ using model $i$ including inference and offloading times. 
$$t_{ij} = it_{i}x_{ij}$$
$$t_{(m+1),j} = (it_{(m+1)}+ot_j+r)x_{(m+1),j}$$

where $r$ is a constant representing the response time. Response times are considered constant because the result of the inference is most likely a vector or a string with a fixed length.

\subsection{Processing Energy}
Let $oe_j$ be the energy cost of offloading job $j$ to the edge server. $oe_j$ depends on the job size $s_j$ and $c$ the energy cost of sending 1 unit of data through the communication channel. $c$ depends on multiple parameter including the communication medium (e.g., Wi-Fi, Cellular or Bluetooth) and its specific configuration (transmission power, reception power, and idle power) taking into account overhead and control data (e.g., headers, acknowledgment packets), in addition to optimization strategies such as data compression and adaptive transmission power based on signal strength. $c$ can be estimated either by externally measuring the power consumption of data transmissions or calculated internally using the communication medium parameters. We prefer the more accurate and simpler method which is to measure the energy consumption externally. We define the offload energy of job $j$ by: 

$$oe_j =  s_jc$$ 

Let $ie_{i}$ be the energy cost of performing the inference of any job using model $i$. The inference energy cost is considered negligible compared to the offloading energy cost and therefore is set to be a very small constant. This constant can be estimated using the inference time and the maximum power consumption of the processor under full load declared by the manufacturer. We define $e_{ij}$ the energy cost of processing job $j$ using model $i$ including inference and offloading energy costs as :

$$e_{ij} = ie_{i}x_{ij}$$

$$e_{(m+1),j} = (ie_{(m+1)}+oe_j)x_{m+1,j}$$

\subsection{An Optimization Problem}
At each time interval the edge node's main objective is to select which models to use for each inference job in order to maximize the total accuracy of each time frame while respecting the given time and energy constraints. Let $x_{ij}=\{0, 1\}$ be a variable where $x_{ij} = 1$ when job $j$ is selected for model $i$ and $x_{ij} = 0$ otherwise. This problem is a maximization problem which can be formulated as follows:

\begin{equation}
\max a = \sum_{j=1}^{n}\sum_{i=1}^{m+1} x_{ij}a_i
\label{eq:maximization}
\end{equation}

Given $E$ and $T$ as the energy and time constraints respectively, Equation \ref{eq:maximization} is subject to:

\begin{equation}
\sum_{j=1}^{n}\sum_{i=1}^{m+1}x_{ij}t_{ij} \leq T
\label{eq:time_constraint}
\end{equation}

\begin{equation}
\sum_{j=1}^{n}\sum_{i=1}^{m+1}x_{ij}e_{ij} \leq E
\label{eq:energy_constraint}
\end{equation}

\begin{equation}
\sum_{i=1}^{m+1}x_{ij} = 1, \forall j \in N
\label{eq:sol_exist}
\end{equation}

where Equation \ref{eq:time_constraint} keeps the selected set of models runtime under the given time constraint. While Equation \ref{eq:energy_constraint} does the same for energy consumption. Equation \ref{eq:sol_exist} guarantees the existence of a solution. 

To analyse the complexity of this problem we draw similarity to the famous knapsack problem (KP). In our case, we are trying to fill our knapsack with inference models (i.e., objects with weights $w_i$ for each dimension $i$) which have weights in 2 dimensions namely inference time and energy cost, therefore the knapsack has 2 weight capacities $W_i$ for each dimension which makes our problem similar to the multi-dimensional KP.  The goal is to try and fill the knapsack with inference models respecting the knapsack capacities and maximizing the profit which in our case is the accuracy. Note that we can reuse models in each selection process therefore we are dealing with the unbounded multi-dimensional variant of the KP (UMdKP).

The UMdKP is classified as strongly NP-hard and is therefore exceptionally difficult to solve optimally. It belongs to a class of problems (NP-hard) for which no efficient solution algorithm is known, and solving it optimally for all instances is believed to require exponential time. As a result, solving this problem often relies on approximation algorithms or heuristic methods that provide near-optimal solutions in a reasonable amount of time.\cite{cacchiani2022knapsack}

The recursive solution for the UMdKP involves solving subproblems in a recursive manner and can be inefficient due to overlapping subproblems and redundant computations. The time complexity of the recursive solution is typically exponential. Let $n$ be the number of items, $m$ be the number of dimensions, and $C$ be the maximum numeric value among the item attributes and knapsack capacities. The recursive solution explores all possible combinations of items, considering both including and excluding each item at each recursive step. The time complexity can be expressed as $O(2^{nmC})$, where each item can be either included or excluded for each dimension, and there are $nm$ dimensions to consider.

This exponential time complexity makes the recursive solution impractical for large instances of the UMDKP, and it becomes inefficient as the problem size increases. To address this issue and improve efficiency, dynamic programming techniques, such as memoization (caching intermediate results), are often employed to avoid redundant computations and reduce the effective time complexity.

The pseudo-polynomial UMDKP typically leverages dynamic programming to efficiently solve instances with relatively small numerical values associated with item attributes and knapsack capacities. The time complexity of the pseudo-polynomial time algorithm is polynomial in the numeric values of the attributes and capacities. The time complexity is often expressed as $O(nmC)$, where $n$ is the number of items, $m$ is the number of dimensions, and $C$ is the maximum numeric value involved.

Because of the limited computing resources onboard the edge nodes and the complexity of the optimization problem under the given time and energy constraints it is impractical to use the recursive or the DP solutions. Therefore, we need to design a fast and lightweight scheme which should be as accurate as the recursive and DP solutions while having minimal energy and runtime footprints. For this reason we look at meta-heuristic methods which are very popular with complex optimization problems specifically Genetic algorithms (GA), Particle Swarm Optimization (PSO) and Ant Colony Optimization (ACO) in the case of the UMdKP.\cite{ezugwu2019comparative} After performing extensive experiments (see Section \ref{sec:exp_results}) we chose GA as our base to design a lightweight selection scheme.

\section{A Lightweight Hybrid Genetic Algorithm Selection Scheme}
\label{sec:lgsto}
In this section we propose a lightweight hybrid genetic algorithm for selective task offloading (LGSTO). In Algorithm \ref{alg:gasto_steps} we present the main steps of LGSTO.

\begin{algorithm}[H]
\caption{Main steps of LGSTO.}
\label{alg:gasto_steps}
\begin{algorithmic}
\STATE $\mathbf{Step 1}$: Initialize population
\STATE $\mathbf{Step 2}$: Evaluate and Rank population fitness
\STATE $\mathbf{Step 3}$: If termination criteria satisfied go to $\mathbf{Step 7}$
\STATE $\mathbf{Step 4}$: Explore neighborhood of the best solution
\STATE $\mathbf{Step 5}$: Produce new generation using Tournament Selection, Crossover and Mutation
\STATE $\mathbf{Step 6}$: Go to $\mathbf{Step 2}$
\STATE $\mathbf{Step 7}$: Return the best solution
\end{algorithmic}
\end{algorithm}

In the following sections we explain each step in detail while justifying our design choices. 

\subsection{Population Initialization}
Considering a population of solutions where each solution is a vector of model indices corresponding to each job. We experimented with multiple initialization schemes including the Latin Hyper-Cube Sampling and Uniform Random Initialization and found that these schemes only add runtime overhead while not providing any improvements to either the accuracy or the convergence speed of the genetic algorithm, therefore in our solution we preferred using a simple Pseudo-random numbers generator based on a modified version of Donald E. Knuth's subtractive random number generator algorithm.\cite{knuth2014art}

Given the population size $p$ we define the set of solutions $S_{i} \in P$ where $P = \{1,..,p\}$ and $S_{i} = \{r_1, r_2, ..,r_n\}$ is a vector of length $n$ which represents the number of inference jobs at any time period while $r$ is the model index assigned to each job. At first the $r$ values are initialized randomly.

$$r_j = Random(1, m) \hspace{0.5cm}  \forall j \in N$$

\subsection{Fitness Evaluation and Ranking}
The population is evaluated against the fitness function $f$ which is in our case the sum of the average accuracies of each model in the solution. The fitness function returns $-\infty$ (in practice $f$ returns the smallest number representable by a double variable) if the sum of average runtimes and the sum of average energy consumptions for a given solution is not within the given constraints represented in Equation \ref{eq:time_constraint} and Equation \ref{eq:energy_constraint} respectively. Finally, the population is sorted in descending order according the fitness values.
\begin{equation}
\label{eq:fitness}
f = \sum a_{i} \hspace{0.5cm} \forall i \in P
\end{equation}

We take advantage of early breaking out of computing Equation \ref{eq:fitness} in the case where a constraint is surpassed before reaching the end of the solution.

\subsection{Termination Criteria}
Our proposed scheme keeps track of the top-1 ranking solution of each generation to compare the best results and determine whether the algorithm has already converged to an optimal solution. This approach speeds up the algorithm and eliminates the unnecessary processing of upcoming generations if the optimal solution is reached in earlier generations. A parameter $TC$ is introduced to specify the number of the most recent best solutions to compare, where if the last $TC$ best solutions are the same we consider the algorithm already converged to an optimal solution. The termination criteria is checked after every $TC$ generations to reduce the computation overhead.

\subsection{Neighborhood Exploration}
After ranking each generation using the fitness function, we propose exploring the neighborhood of the best solution of each generation $S_0$ to look for better nearby solutions. We define $W$ a walk distance for searching the neighboring solutions. We take each model index of $S_0$ and increase/decrease it within the index limits by the walking distance $W$ while evaluating each neighbor against $S_0$, if we find neighbors with a higher fitness value we add it to neighbors list and later to the new generation. We define the searching scheme in Algorithm \ref{alg:neighborhood_exploration}.

\begin{algorithm}[H]
\caption{Neighborhood Exploration} 
\label{alg:neighborhood_exploration}
\begin{algorithmic}
\STATE FindBestNeighbors($S_0$, $W$):
\STATE $neighbors \xleftarrow{} \{\}$
\STATE for each $r_j \in S_0$:
\STATE \hspace{0.5cm}for each $k=1$ to $W$:
\STATE \hspace{1cm}for each $d$ in $\{1, -1\}$:
\STATE \hspace{1.5cm}$r_j = r_j + (k\times d)$
\STATE \hspace{1.5cm}if $f(S) > f(S_0)$: 
\STATE \hspace{2cm}$neighbors = neighbors + \{S\}$
\end{algorithmic}
\end{algorithm}

\subsection{Reproduction Process}
To create the new generation we chose a mating scheme in which we select parents from the population using tournament selection. We select a subset of the population of size $ts$ and we perform a tournament in which we compare the selected solutions and find the fittest solution. We perform this tournament selection twice to find both parents and then we perform the crossover. We tested  1-point, 2-point and discrete uniform crossover (DUC) methods and found that the latter performed the best compared to the first two methods. In DUC we create the offspring by randomly selecting a model index $r$ from each parent with a $50\%$ probability.

The newly generated offspring undergo probabilistic mutations to keep the population diverse and prevent it from converging to a local optimal solution. A mutation probability parameter is introduced to determine whether an offspring undergoes a mutation.  A higher mutation probability increases the chances of introducing more random changes, which can be beneficial for exploration but may also disrupt promising solutions. Conversely, a lower mutation probability might lead to slower exploration but can help preserve promising solutions.

We introduce a fading parameter to the mutation probability which allows the adjustment of the mutation probability during the evolution process dynamically. Fading mutation probability involves reducing the mutation rate as the algorithm progresses through generations. The goal is to initially promote exploration with a higher mutation rate to discover diverse regions of the solution space and gradually decrease it to allow for exploitation and refinement of promising solutions.

The fading mutation probability strategy aims to strike a balance between exploration and exploitation over the course of the algorithm. Higher mutation rates early in the optimization process help the algorithm explore a broad range of solutions, and as the algorithm converges, the mutation rate is reduced to allow for fine-tuning around more promising regions.

We found that linearly fading the mutation probability with a constant parameter accelerated the convergence of the evolution process and allowed the development of optimal solutions in later generations without being disrupted by random mutations.

\begin{algorithm}[H]
\caption{Reproduction Process} 
\label{alg:reproduction}
\begin{algorithmic}
\STATE NewGeneration $\xleftarrow{} \{\}$
\STATE for $i = 0$ to $PopulationSize$:
\STATE \hspace{0.5cm} $p_1 = TournamentSelection(Population)$
\STATE \hspace{0.5cm} $p_2 = TournamentSelection(Population)$
\STATE \hspace{0.5cm} $(o_1, o_2) = DiscreteUniformCrossover(p_1, p_2)$
\STATE \hspace{0.5cm} if ($MutationProbability > 0$):
\STATE \hspace{1cm} $Mutate(o_1, MutationProbability)$
\STATE \hspace{1cm} $Mutate(o_2, MutationProbability)$
\STATE \hspace{0.5cm} if ($f(o_1) > 0$):
\STATE \hspace{1cm} $NewGeneration = NewGeneration + \{o_1\}$
\STATE \hspace{0.5cm} if ($f(o_2) > 0$):
\STATE \hspace{1cm} $NewGeneration = NewGeneration + \{o_2\}$

\STATE $Population \xleftarrow{} NewGeneration$
\STATE $MutationProbability = MutationProbability - FadingParameter$
\end{algorithmic}
\end{algorithm}

\section{Experiments and Results}
\label{sec:exp_results}
In this section, we present the setup used in our experiments followed by a description of the parameters used for tuning LGSTO, estimating the communication times and energy consumption.  Finally, we show the obtained results in comparison to other methods.

\subsection{Setup Configuration}
We conduct our experiments using a Raspberry 4 as an edge node connected via a LAN connection to an edge server represented by a more powerful computer with dedicated graphics for hardware accelerated inference.

The experiment consists of performing object classification over a stream of images. Using the Imagenet-mini dataset \cite{vinyals2017matching} as a source of data we chose three lightweight object classification models namely: resnet18 and resnet34 \cite{he2015deep} and shufflenetv2 \cite{ma2018shufflenet} deployed in the edge node. In the edge server we deploy resnext101 \cite{xie2017aggregated} which is a larger and more accurate model. During the deployment phase we run tests on these models to estimate the average runtime and energy consumption of each inference model (see Table \ref{tab:models_averages}).

\begin{table}[!t]
\caption{Models average accuracies and inference times\label{tab:models_averages}}
\centering
\begin{tabular}{|c||c||c|}
\hline
   & \begin{tabular}[c]{@{}l@{}}Average\\ Accuracy\\(\%)\end{tabular}
   & \begin{tabular}[c]{@{}l@{}}Average\\ Inference\\ Time\\ (ms)\end{tabular} \\
  \hline
resnet18 & 72.01986328 & 28.07417981 \\
  \hline
resnet34 & 76.79044298 & 42.45949233 \\
  \hline
shufflenetv2 & 66.15977267 & 19.44331129 \\
  \hline
\begin{tabular}[c]{@{}l@{}}resnext101\\ (Edge Server)\end{tabular} & 87.05788745 & 5.1610317 \\ 
  \hline
\end{tabular}
\end{table}

\subsection{Experiment Parameters}
In this section we present the parameters used for performing experiments including the LGSTO parameters which where used to fine tune the proposed scheme. In addition to the experiment parameters and constraints. Finally, the network setup parameters are presented.

\subsubsection{LGSTO Parameters}
The process of obtaining the best LGSTO parameters is automated using scripts to find the best values striking a balance between performance and accuracy. This process can be performed on device during the deployment phase. The parameters are presented in Table \ref{tab:LGSTO_params}.

\begin{table}[!t]
\caption{LGSTO Parameters\label{tab:LGSTO_params}}
\centering
\begin{tabular}{|c||c|}
\hline
  Mutation Probability & 0.3 \\ 
  \hline
  Fading Factor & 0.01 \\ 
  \hline
  Termination Count & 3  \\ 
  \hline
  Population Size & 100 \\
  \hline
  Generations Count & 200 \\
  \hline
  Tournament Size & 20 \\
  \hline
\end{tabular}
\end{table}

\subsubsection{Experiment Parameters}
We selected an arbitrary logical set of time and energy constraints for a typical edge device such as a smartphone (see Table \ref{tab:exp_params}). Please note that we are using the term energy constraint typically measured in KWh, however in our context it is more convenient to use power consumption measured in watts. Therefore in the following sections we will be only using the power consumption instead of energy. The number of jobs is the number of entries in the Imagenet-mini dataset. The jobs per time slot variable along side the Time constraint are selected to represent a real-time video processing application which produces approximately 30 processed frames per second.

\begin{table}[!t]
\caption{Experiment Parameters\label{tab:exp_params}}
\centering
\begin{tabular}{|c||c|}
\hline
Time Constraint (ms) & 350 \\
\hline
Energy Constraint (W) & 100 \\
\hline
Jobs Count & 3923 \\ 
\hline
Jobs Per Time Slot & 10\\
\hline
\end{tabular}
\end{table}

\subsubsection{Environment Parameters}
To obtain a good approximation for the power consumption of data transmission using Ethernet we can either measure it externally or refer to the manufacturer's data sheets. In our case, we refer to the Texas Instruments Ethernet power consumption for a 100 Base-TX Full Duplex Operating Mode connection for an Ethernet chip.\cite{texas2013} We considered the worst case and found the values shown in Table \ref{tab:env_params}.

\begin{table}[!t]
\caption{Environment Parameters\label{tab:env_params}}
\centering
\begin{tabular}{|c||c|}
\hline
Data Rate & 100 Mbps \\
\hline
Power Cost per Megabyte & 1.8 w \\
\hline
\end{tabular}
\end{table}

\subsection{Implementation of comparable schemes}
Since no similar work in the literature has considered multiple inference models with both time and energy constraints, we compare our results to other schemes, such as the Naive approach and Dynamic Programming (DP) for pseudo-polynomial solutions. In addition to Non-dominated Sorting Genetic Algorithm II (NSGA-II) \cite{deb2002}, Particle Swarm Optimization (PSO), Ant Colony Optimization (ACO) and a standard genetic algorithm (GA) for the evolutionary schemes. A description along with the configuration parameters for each scheme is presented in the following section. Note that for finding the best parameters for each scheme we employed the same automated method as LGSTO.

The Naive (Brute-force) scheme has been optimized using memoization with a time complexity of $O(M\times T \times E \times N)$ and space complexity of $O(M\times T \times E \times N)$. While the Dynamic Programming scheme has the same time complexity and space complexity of $O(T \times E \times N)$.

NSGA-II is implemented according to the proposed work in \cite{deb2002} while setting the Population Size to $100$ and Max Generations Count to $10$ with a Mutation probability of $0.3$.

The PSO scheme is implemented with a swarm size of $2000$ particles and Max iterations is set to $30$. The initial velocities are set to $\{1, -1\}$.

The ACO scheme is implemented with $200$ ants and an Evaporation Rate of $0.1$ and Max iterations is set to $50$.

The Genetic Algorithm scheme is implemented using two reproduction methods namely a  gene pool (referred to as GA-GP) and 1-point crossover (referred to as GA-CR). All other parameters are similar to LGSTO.

\subsection{Results}
\begin{figure}[!t]
\centering
\includegraphics[width=3.4in]{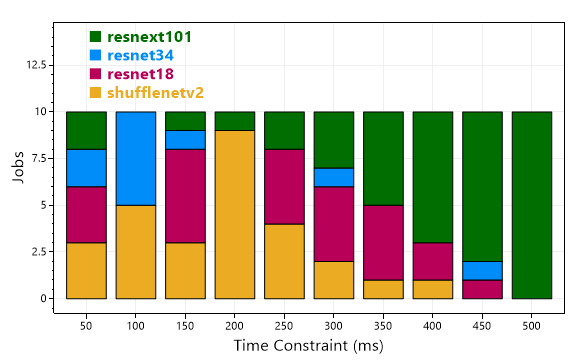}
\caption{Model allocation for 10 inference jobs with varying time constraint and power constraint of 20 (w) using LGSTO.}
\label{fig:vary_time}
\end{figure}

We start by first examining the model allocation results of LGSTO under varying time constraints with a fixed power constraint at 20 (w) as shown in Fig \ref{fig:vary_time}. We observe that with time constraints lower than 200 (ms) the edge server model (i.e. resnext101) is rarely used, however, as the time constraint increases above 200 (ms) it is used more and more to maximize accuracy and take advantage of the additional time.   

\begin{figure}[!t]
\centering
\includegraphics[width=3.4in]{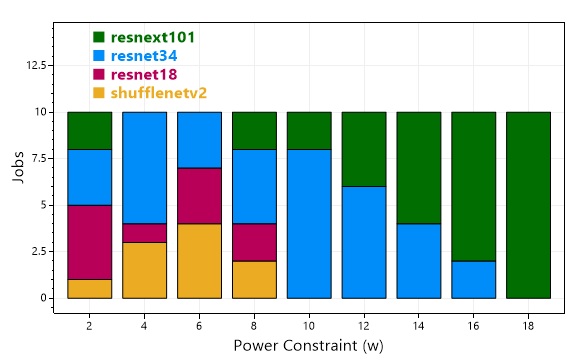}
\caption{Model allocation for 10 inference jobs with varying power constraint and time constraint of 500 (ms) using LGSTO.}
\label{fig:vary_pow}
\end{figure}

Similarly, we examine LGSTO results when varying the power constraint and fixing the time constraint at 500 (ms) as shown in Fig \ref{fig:vary_pow}. We observe a similar behavior to when we varied the time constraint however, in addition to the edge server model (i.e. resnext101) we see the second preferred model that is the resnet34 which has a higher accuracy and inference times and thus higher inference power consumption (see Table \ref{tab:models_averages}) this behavior is a result of LGSTO taking advantage of the higher available power budget.

\begin{figure}[!t]
\centering
\includegraphics[width=3.4in]{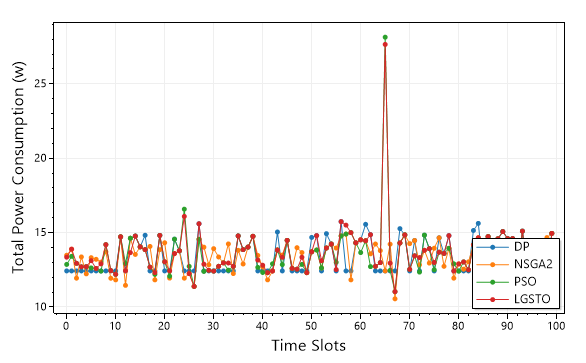}
\caption{Total power consumption for 100 time slots.}
\label{fig:total_power}
\end{figure}

\begin{figure}[!t]
\centering
\includegraphics[width=3.4in]{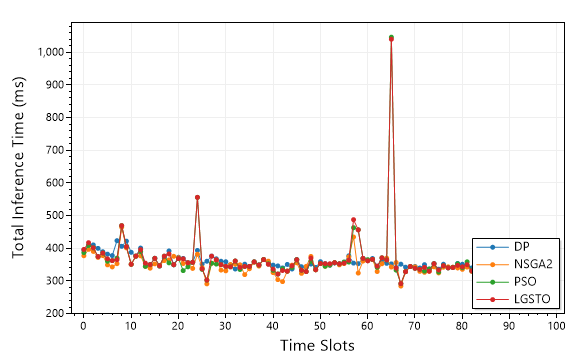}
\caption{Total inference time for 100 time slots.}
\label{fig:total_inf}
\end{figure}

By analysing the total power consumption for each time slot for 100 time slots shown in Fig \ref{fig:total_power} we can see all schemes are producing schedules that consume slightly higher power than the given constraint which is expected since the schemes rely on an estimated average power consumption value to assign models. We also notice a few spikes in power consumption which is explained by large size images deviating from the average size in the dataset. Similarly, in Fig \ref{fig:total_inf} we can see the time constraint is mostly respected at 350ms except for a few spikes for the same reason as the power consumption graph.

\begin{figure}[!t]
\centering
\includegraphics[width=3.4in]{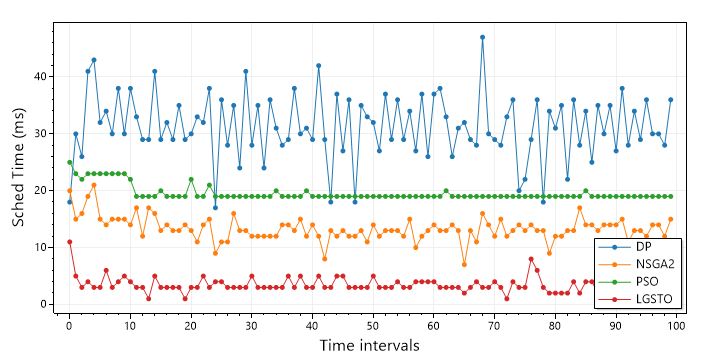}
\caption{Scheduling times for 100 time slots.}
\label{fig:sched_times}
\end{figure}

We measure the scheduling time representing the time each scheme takes to assign models to jobs. Fig \ref{fig:sched_times} shows scheduling times for 100 time slots. We observe that LGSTO has the lowest scheduling time at under 5ms compared to DP, NSGA-II and PSO. We also see that DP and thus pseudo-polynomial time schemes have a larger amount of variation in scheduling times depending on different problem parameters which is not suitable for real-time applications. On the other hand, evolutionary schemes are unaffected by problem variations and always produce solutions at a stable time which is favorable for real-time applications.

\begin{figure}[!t]
\centering
\includegraphics[width=3.4in]{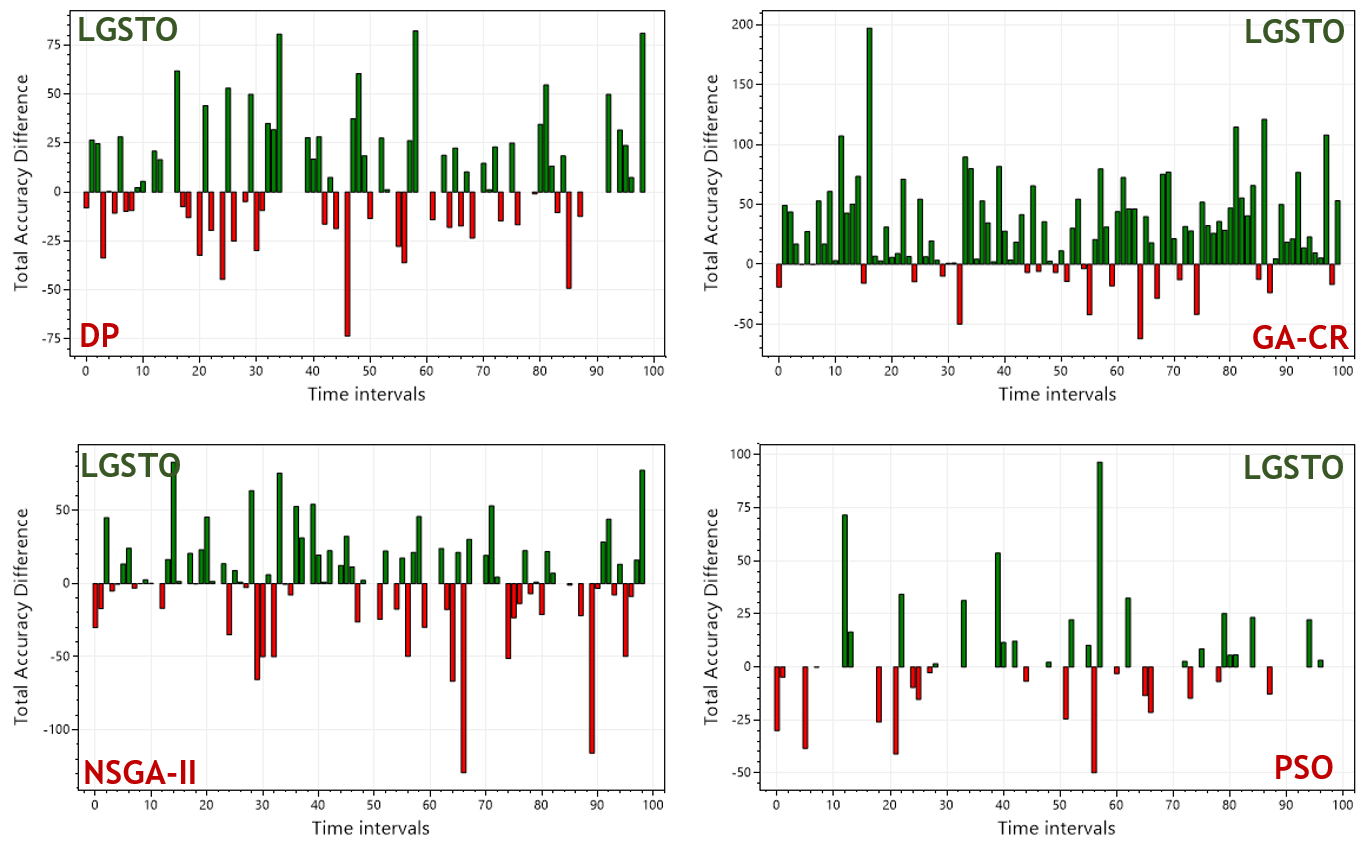}
\caption{Total Accuracy differences between LGSTO and other schemes for 100 time slots.}
\label{fig:acc_diffs}
\end{figure}

Comparing the total accuracy difference of LGSTO with other schemes as shown in Fig \ref{fig:acc_diffs} for 100 time slots we see that LGSTO produces schedules that result in higher overall accuracy per time slot compared to all other schemes. The next best scheme is PSO producing almost similar accuracy however at a significantly higher scheduling time.

\begin{figure}[!t]
\centering
\includegraphics[width=3.4in]{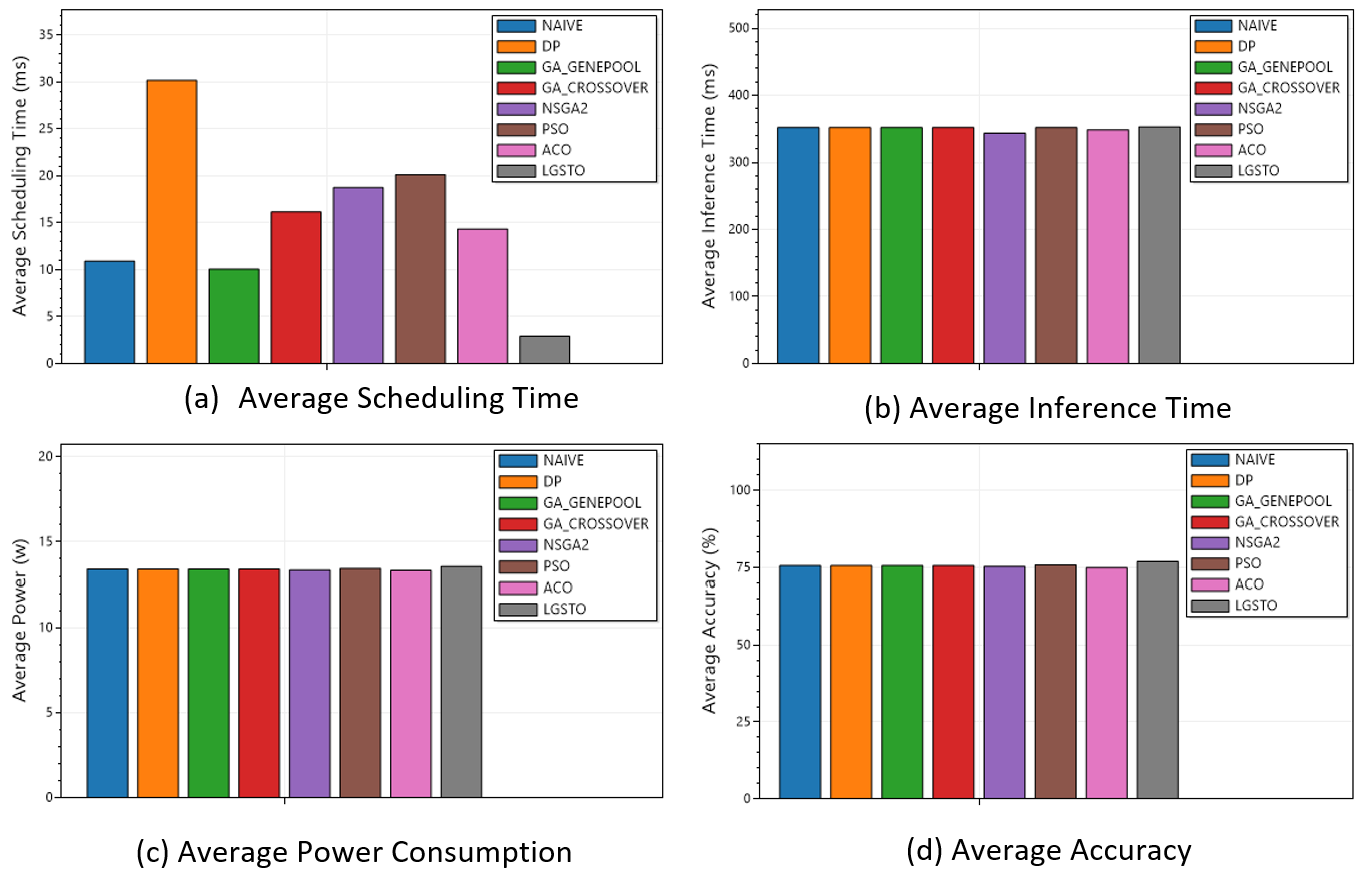}
\caption{Experiment average metrics for the Imagenet-mini dataset.}
\label{fig:all_averages}
\end{figure}

To summarize all the results, We calculate the total inference time and power consumption for each time slot and then average out all the totals across all time slots. The average accuracy is measured by calculating the average for each time slot and then average out all accuracies across all time slots. Finally, the scheduling time is averaged out across all time slots. The results of our experiments on the Imagenet-mini dataset are presented in Table \ref{tab:avg_results} and illustrated in Fig \ref{fig:all_averages}.

Looking at the results in Table \ref{tab:avg_results} we can see that LGSTO is producing schedules that give on average higher accuracy relative to the other schemes while having the lowest scheduling time which results in the lowest total processing time which is the closest to the given time constraint for a time-slot.
All schemes are producing schedules with similar average power consumption and average inference times apart from NSGA-II and ACO with lower inference times which has incidentally reduced their average accuracy as a result of not taking advantage of higher accuracy models with higher inference times.
The pseudo-polynomial schemes produce identical scheduling results with identical averages apart from the scheduling times in which we see that the Naive approach with memoization is performing better than the DP scheme. 

\begin{table}[!t]
\caption{Averages over all time steps\label{tab:avg_results}}
\centering
\begin{tabular}{|c||c||c||c||c||c|}
\hline
 & 
 \begin{tabular}[c]{@{}l@{}}
 Average\\ Accuracy\\(\%)
 \end{tabular} 
 &  
 \begin{tabular}[c]{@{}l@{}}
 Average \\Power \\cons-\\umption\\(w)\end{tabular} 
 & 
 \begin{tabular}[c]{@{}l@{}}
 Average\\ Inference\\ Time\\ (ms)\end{tabular}
 & 
 \begin{tabular}[c]{@{}l@{}}
 Average\\ Sche-\\duling\\ time\\ (ms)\end{tabular} 
 &
 \begin{tabular}[c]{@{}l@{}}
 Total \\time\\(ms)
 \end{tabular} \\
\hline
NAIVE & 75.58 & 13.40 & 352.02 & 10.92 & 362.94 \\
\hline
DP & 75.58 & 13.40 & 352.02 & 30.17 & 382.20 \\
\hline
GA-GP & 75.59 & 13.39 & 352.01 & 10.07 & 362.08\\
\hline
GA-CR & 75.59 & 13.40 & 351.97 & 16.16 & 368.13\\
\hline
NSGA2 & 75.30 & 13.34 & 343.57 & 18.75 & 362.32 \\
\hline
PSO & 75.80 & 13.44 & 352.15 & 20.13 & 372.28\\
\hline
ACO & 74.90 & 13.32 & 348.72 & 14.34 & 363.07\\
\hline
LGSTO & 76.05 & 13.55 & 352.06 & 2.91 & 354.97\\
\hline
\end{tabular}
\end{table}

\section{Conclusion}
\label{sec:conclusion}
In this work we studied the selective inference task offloading problem in edge nodes under time and energy constraints. In which the edge nodes are equipped with multiple local inference models varying in size and thus in accuracy and power requirement. In addition to an edge server equipped with a more powerful inference model which edge nodes can offload inference jobs to. We analyse the problem of selecting the appropriate inference model for a given inference job under time and energy constraints while maximizing accuracy where we demonstrate that this problem is an instance of the known unbounded multidimensional knapsack problem which is considered a strongly NP-hard problem. Therefore, we propose LGSTO a light weight hybrid genetic algorithm to solve this problem. We perform experiments
on the Imagenet-mini dataset and compare our scheme against classic genetic algorithms from the literature in addition to pseudo-polynomial time methods namely Dynamic programming and the naive method optimized with memoization. Other evolutionary methods are also compared such as Particle Swarm Optimization and Ant colony Optimization. Results show that LGSTO performed $70\%$ faster than the best other schemes while producing schedules with higher average accuracy. Therefore, LGSTO is considered suitable for real-time applications with energy constrained edge devices.

Although this work has mainly focused on solving the inference model selection problem under time and energy constraints, we only considered a wired communication channel between the edge node and edge server which results in a predictable offload times whereas wireless channels depend more on environment changes and require more sophisticated approximation methods. Another aspect that has not been considered in this work is the parallel execution of inference jobs between edge nodes and edge servers. These limitations are being considered for a future work.

\section*{Acknowledgments}
This article was funded by National Natural Science Foundation of China
61872038.



\bibliographystyle{IEEEtran}
\bibliography{refs}

\newpage


\vskip -3\baselineskip 

\begin{IEEEbiography}[{\includegraphics[width=1in,height=1.25in,clip,keepaspectratio]{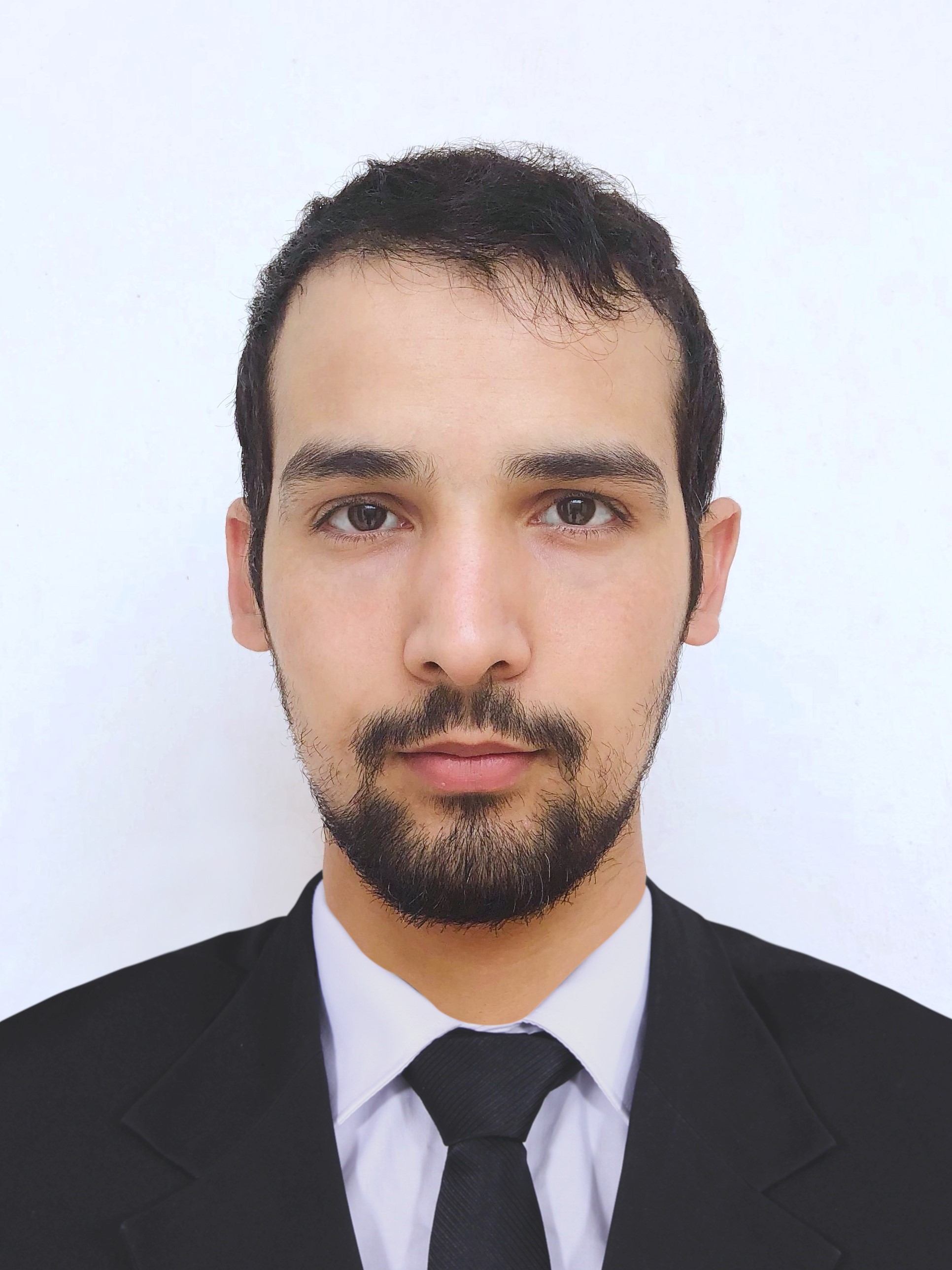}}]{Abdelkarim Ben Sada}
Received his BSc in Computer Science in 2014 from the University of Djelfa Algeria, and his MSc degree in 2016 majoring in Networking and Distributed Systems from the University of Laghouat Algeria. He is currently pursuing his PhD degree at the University of Science and Technology Beijing China. His research interests include Computer Vision, Machine Learning and Internet of Things.
\end{IEEEbiography}

\vskip -2\baselineskip 
\begin{IEEEbiography}[{\includegraphics[width=1in,height=1.25in,clip,keepaspectratio]{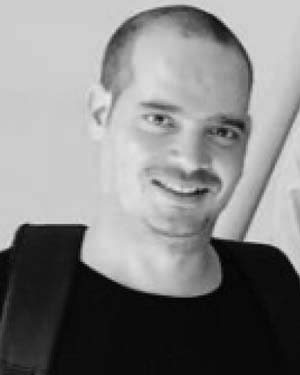}}]{Amar Khelloufi}
Received the B.S. degree (Hons.) in computer science from the Faculty of Sciences and Technology, Ziane Achour University of Djelfa, Djelfa, Algeria, in 2012, and the M.S. degree in distributed information systems from the Faculty of Sciences, University of Boumerdès, Boumerdès, Algeria, in 2014. He is currently pursuing the Ph.D. degree with the School of Computer and Communication Engineering, University of Science and Technology Beijing, Beijing, China. His current research focuses on Internet of Things, blockchain applications, edge computing, and distributed systems.
\end{IEEEbiography}

\begin{IEEEbiography}[{\includegraphics[width=1in,height=1.25in,clip,keepaspectratio]{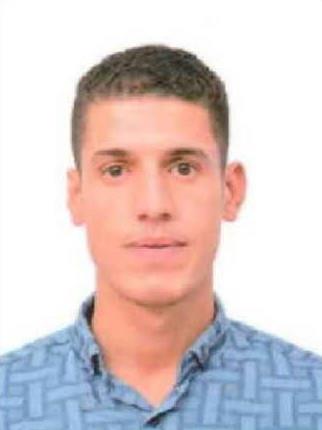}}]{Abdenacer Naouri}
He is currently a Ph.D. candidate at the University of Science and Technology Beijing China, Beijing, China. He 
received his B.S. degree in computer science from the University of Djelfa Algeria, in 2011, and the  M.Sc. degree in networking and distributed systems from the University of Laghouat Algeria, Laghouat, Algeria, in 2016. His current research interests include Cloud computing, Smart communication, machine learning, Internet of vehicles  and Internet of Things. \end{IEEEbiography}

\begin{IEEEbiography}[{\includegraphics[width=1in,height=1.25in,clip,keepaspectratio]{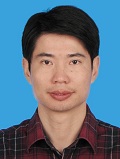}}]{Huansheng Ning}
Received his B.S. degree from Anhui University in 1996 and his Ph.D. degree from Beihang University in 2001. Now, he is a professor and vice dean of the School of Computer and Communication Engineering, University of Science and Technology Beijing, China. His current research focuses on the Internet of Things and general cyberspace. He has presided many research projects including Natural Science Foundation of China, National High Technology Research and Development Program of China (863 Project). He has published more than 150 journal/conference papers, and authored 5 books. He serves as an associate editor of IEEE Systems Journal (2013-Now), IEEE Internet of Things Journal (2014-2018), and as steering committee member of IEEE Internet of Things Journal (2016-Now).
\end{IEEEbiography}

\begin{IEEEbiography}[{\includegraphics[width=1in,height=1.25in,clip,keepaspectratio]{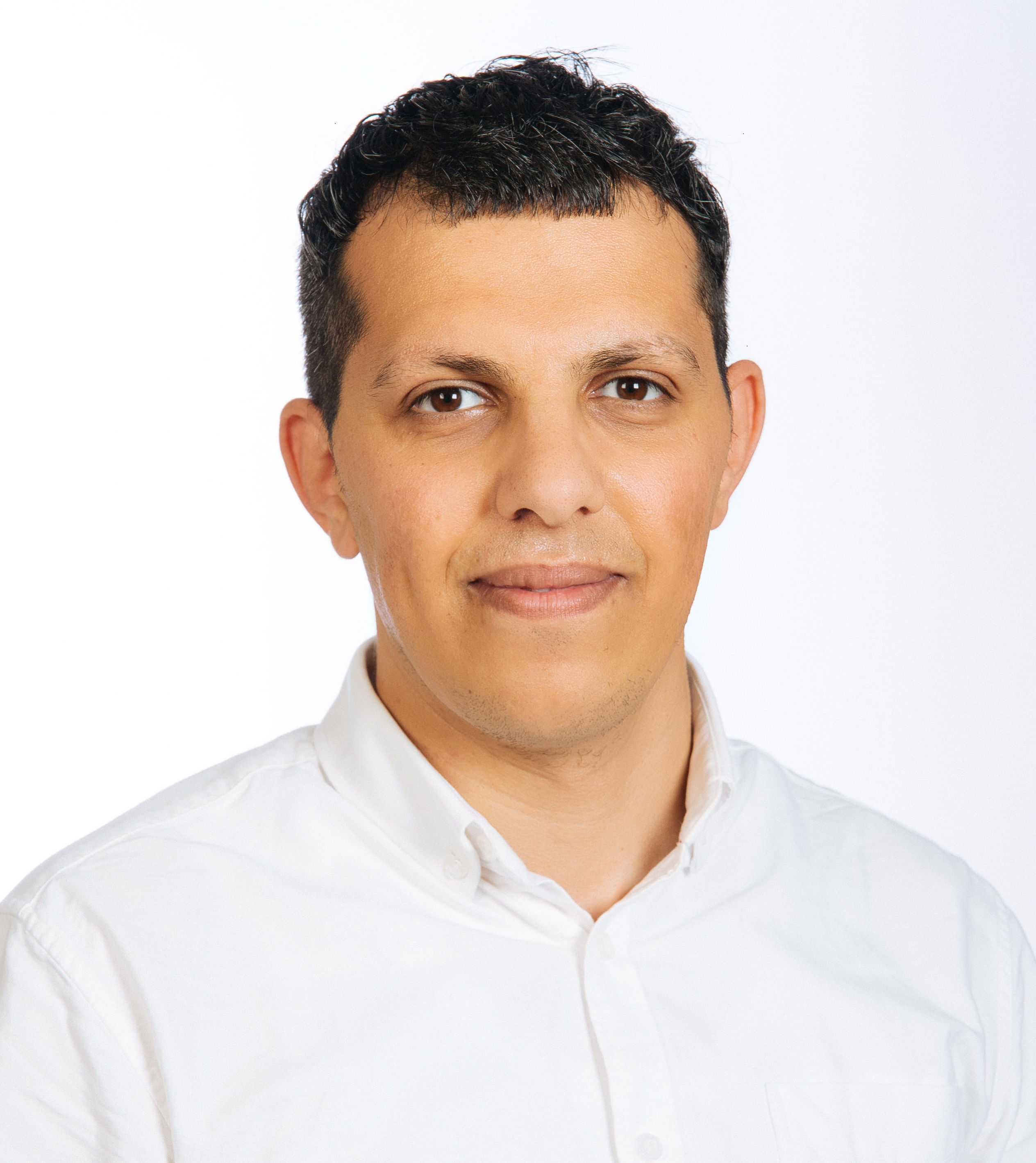}}]{Sahraoui Dhelim} is a senior postdoctoral researcher at University College Dublin, Ireland. He was a visiting researcher at Ulster University, UK (2020-2021). He obtained his PhD degree in Computer Science and Technology from the University of Science and Technology Beijing, China, in 2020. And a Master's degree in Networking and Distributed Systems from the University of Laghouat, Algeria, in 2014. He serves as a guest editor in several reputable journals, including Electronics journal and Applied Science Journal. His research interests include Social Computing, Smart Agriculture, Deep-learning, Recommendation Systems and Intelligent Transportation Systems.
\end{IEEEbiography}

\vfill

\end{document}